\documentclass[a4paper,conference]{IEEEtran}
\usepackage[utf8]{inputenc}
\usepackage[T1]{fontenc} 
\IEEEoverridecommandlockouts

\usepackage{cite}
\usepackage{amsmath,amssymb,amsfonts}
\usepackage{graphicx}
\usepackage{textcomp}
\usepackage{xcolor}
\usepackage[english]{babel}
\usepackage[letterpaper,top=2cm,bottom=2cm,left=2cm,right=2cm,marginparwidth=1.75cm]{geometry}
\usepackage[hyphens]{url}
\usepackage[hidelinks=true]{hyperref}
\usepackage[subpreambles=false]{standalone}
\usepackage[labelfont=footnotesize, textfont=footnotesize, figurename=Fig., tablename=Tab.]{caption} 
\usepackage{cleveref}
\usepackage{textcomp}
\usepackage{longtable}
\usepackage{flushend}
\usepackage{pgfplots}
\usepackage{pifont}
\usepackage[nohyperlinks, nolist]{acronym}
\def\BibTeX{{\rm B\kern-.05em{\sc i\kern-.025em b}\kern-.08em
    T\kern-.1667em\lower.7ex\hbox{E}\kern-.125emX}}

\makeatletter 
\newcommand{\linebreakand}{%
  \end{@IEEEauthorhalign}
  \hfill\mbox{}\par
  \mbox{}\hfill\begin{@IEEEauthorhalign}
}
\makeatother 

\usepackage{booktabs}
\usepackage{float} 

\usepackage{subfig}
\usepackage{algorithm}
\usepackage{algpseudocode}
\usepgfplotslibrary{groupplots} 

\usepackage{siunitx}

\usepackage{tikz}
\usepackage{wasysym}
\usepackage{makecell}
\usepackage{adjustbox}
\usepackage{tikz-3dplot}
\usetikzlibrary{automata}
\usetikzlibrary{positioning}
\usetikzlibrary{calc}
\usetikzlibrary{fit}
\usetikzlibrary{spy}
\usetikzlibrary{arrows.meta}
\usetikzlibrary{backgrounds}
\usetikzlibrary{plotmarks}
\usetikzlibrary{shapes,shapes.geometric,shapes.symbols,shapes.arrows, shapes.multipart}
\usetikzlibrary{patterns}

\newcommand{\h}{3cm}
\newcommand{\w}{3cm}
\newcommand{\hd}{0.2cm}
\newcommand{\vd}{-1.75cm}
\newcommand{\marsiz}{1pt}
\newcommand{\mar}{o}
\definecolor{clr1}{HTML}{00FFFF}
\definecolor{clr2}{RGB}{110,110,110}
\newcommand{\cmark}{\ding{51}}%
\newcommand{\xmark}{\ding{55}}%
\newcommand{\xx}{5.3}
\newcommand{\xxx}{10.7}
\newcommand{\xxxx}{11.3}
\newcommand{\xxxxx}{11}
\newcommand{\yy}{5.1}
\newcommand{\yyy}{4.8}
\newcommand{\yyyy}{4.2}
\newcommand{\podelta}{0.03}
\newcommand{\ahwidth}{0.5}

\begin{document}

\title{
Realtime-Capable Hybrid Spiking Neural Networks for Neural Decoding of Cortical Activity

\thanks{
The project on which this report is based was sponsored by the German Federal Ministry of Education and Research under grant number 16ME0801. The responsibility for the content of this publication lies with the author.
}
}

\author{\IEEEauthorblockN{Jann Krausse$^*$,\!$^{1,2}$ Alexandru Vasilache$^*$,\!$^{1,3}$  \\ Klaus Knobloch,\!$^{2}$ Juergen Becker \!$^{1}$ \\
    }
    \IEEEauthorblockA{
    \textit{$^{1}$ Karlsruhe Institute of Technology, Karlsruhe, Germany}\\
    \textit{$^{2}$ Infineon Technologies, Dresden, Germany} \\
    \textit{$^{3}$ FZI Research Center for Information Technology, Karlsruhe, Germany} \\
    }
}

\maketitle
\def\thefootnote{*}\footnotetext{These authors contributed equally to this work.}\def\thefootnote{\arabic{footnote}}

\begin{acronym}[]
    \acro{ann}[ANN]{artificial neural network}
    \acro{snn}[SNN]{spiking neural network}
    \acro{ai}[AI]{artificial intelligence}
    \acro{ibmi}[iBMI]{intra-cortical brain-machine interface}
    \acro{bmi}[BMI]{brain machine interface}
    \acro{gru}[GRU]{gated recurrent unit}
    \acro{lif}[LIF]{leaky integrate-and-fire}
    \acro{sgru}[sGRU]{spiking GRU}
\end{acronym}

\begin{abstract}
\textbf{
\Acp{ibmi} present a promising solution to restoring and decoding brain activity lost due to injury.
However, patients with such neuroprosthetics suffer from permanent skull openings resulting from the devices' bulky wiring.
This drives the development of wireless \acp{ibmi}, which demand low power consumption and small device footprint.
Most recently, \acp{snn} have been researched as potential candidates for low-power neural decoding.
In this work, we present the next step of utilizing \acp{snn} for such tasks, building on the recently published results of the 2024 Grand Challenge on Neural Decoding Challenge for Motor Control of non-Human Primates.
We optimize our model architecture to exceed the existing state of the art on the Primate Reaching dataset while maintaining similar resource demand through various compression techniques.
We further focus on implementing a realtime-capable version of the model and discuss the implications of this architecture.
With this, we advance one step towards latency-free decoding of cortical spike trains using neuromorphic technology, ultimately improving the lives of millions of paralyzed patients.
}
\end{abstract}
\section{Introduction}
Loss of muscle control due to paralysis affects tens of millions of individuals worldwide \cite{paralysis_us, spinalcordinjury_who}. For them, \acp{bmi} that decode brain activity to control external prostheses could pose a life-changing prospect \cite{bmi, bmi_appl}. However, current so-called \acp{ibmi} suffer from interfacing the brain through a skull opening, necessitating bulky wiring for connectivity \cite{wireless_ibmi, infection_risk}. Critically, this raises the risk of infection and further impairs head mobility, motivating the development of wireless \acp{ibmi} \cite{wireless_ibmi, wireless_ibmi2}. \acp{ibmi} are fully implanted into the skull by enabling wireless communication with the prosthesis control. Nevertheless, with this arise new challenges. Invasive implants demand minimal heat dissipation and have a restricted battery lifetime \cite{thermal_dissipation}. This limits the device's bandwidth, requiring high-quality compression and high energy efficiency.

\Acp{snn} have been proposed as promising candidates for such neural decoders \cite{rel_work_dethier, rel_work_liao}. While neural networks display exceptional decoding abilities, running event-based networks on dedicated neuromorphic hardware facilitates unparalleled low-power execution \cite{ai_bmi, ai_movement_decoding, neuromorphics}. Additionally, \acp{snn} naturally work on event-based data, matching brain data processing. Recently, the 2024 Grand Challenge on Neural Decoding for Motor Control of Nonhuman Primates tasked participants to train novel \acp{snn}-based decoders, leading to the presentation of various promising approaches \cite{neurobench_challenge}. Partaking teams were provided with a set of recordings from the "Nonhuman Primate Reaching with Multichannel Sensorimotor Cortex Electrophysiology" dataset and challenged to optimize for accuracy ($R^2$) or co-optimize for accuracy and resource management, including memory footprint and number of operations \cite{primate_reaching_dataset}.

In this work, we will assess the continuation of our work, which was first presented as one of the winners of the Neural Decoding Challenge \cite{alex_jann}. We introduced a hybrid spiking neural network comprising a temporal convolution and processing through recurrently connected \ac{lif} units. Here, we present the advances regarding our networks by undertaking rigorous hyperparameter optimization and adopting successful methods presented by other teams \cite{zenke_decoding, ini_challenge}. We further apply several compression techniques to reduce model footprint and computational demand. 
Finally, we introduce a version of our model specifically targeting real-time execution and will discuss the implications of such models with non-noticeable latency. We identify this as crucial for the eventual real-world application of our approach. Lastly, we compare our results with the existing state of the art and discuss our findings and the limitations of this work.
\section{Related Work}
The NeuroBench framework and the Primate Reaching task for \acp{snn} have been introduced in \cite{neurobench}. They give a baseline for networks optimized for accuracy and for networks co-optimized for accuracy, memory footprint, and computational demand. Additionally, NeuroBench was used for fair benchmarking of all networks submitted to the Neural Decoding Challenge and is used in our work for that purpose.

The authors of \cite{zenke_decoding} facilitate \acp{snn} with explicit recurrent connections (RSNN) for solving the task given by the Neural Decoding Challenge. They present a large network with a performance far surpassing the baseline given in \cite{neurobench} and a small network, which is a compressed version of the larger one, also beating the respective baseline.

The model architecture used in this work was first presented in \cite{alex_jann}. Like the RSNN, this model surpassed the baselines given by \cite{neurobench} for both the accuracy and co-optimization tracks. Here, the more efficient model is compressed by decreasing the number of layers and their sizes at a substantial cost in terms of $R^2$ decrease.


Although that topic is briefly touched upon by \cite{alex_jann}, we identify a lack of in-depth discussions regarding the real-time execution of the networks in addition to the missing presentation of a respective implementation. Hence, we consider the real-time implementation to be one of the main contributions of this work.
\section{Neural Decoding Task}
\begin{figure}
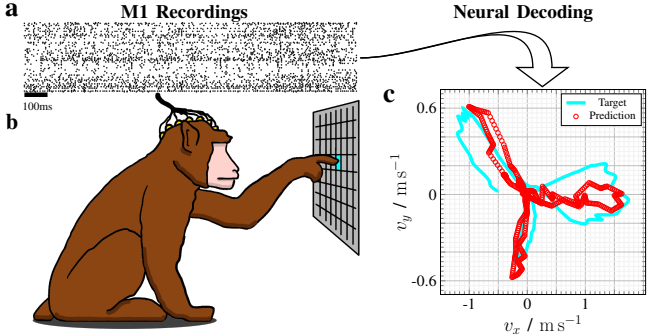

    \centering
    \includestandalone[width=\linewidth]{figures/tikz/primatereaching_new}
    \caption{Visualization of the Primate Reaching dataset and the related neural decoding task. \textbf{b} illustrates the primates reaching for a target while their brain activity is recorded. \textbf{a} shows some resulting spike trains from the motor cortex recordings. \textbf{c} presents the prediction of our \acp{snn} for one of the sequences within the recordings and the respective target.}
    \label{fig:primate_reaching}
\end{figure}

The “Nonhuman Primate Reaching with Multichannel Sensorimotor Cortex Electrophysiology” dataset is a collection of spike recordings from Macaque monkey brains while reaching for targets on an 8x8 grid using one of their arms \cite{primate_reaching_dataset}. The subjects are two different monkeys: Indy, whose left hemisphere was recorded while reaching with the right arm, and Loco, whose right hemisphere was recorded while reaching with the left arm. The primary motor cortex (M1) was recorded in most sessions, comprising 96 input channels. In addition to the 96 M1 channels, the somatosensory cortex (S1) was recorded for the rest of the sessions, adding another 96 channels to the input. The task is to decode the spike recordings into finger-tip velocities of the arm movements projected on the two-dimensional plane the targets were placed on. Figure \ref{fig:primate_reaching} illustrates the task and data. 

The Neural Decoding Challenge stipulated six specific recordings for evaluation of the decoders \cite{neurobench_challenge}. The first three files are recordings of Indy, consisting of M1 data. The latter three are recordings of Loco, comprising spikes from the M1 and S1. We will address the files as I1 to I3 for Indy and L1 to L3 for Loco.
\section{Methods}
\subsection{Network Architecture}
\label{sec:architecture}
The models used in this work are based on the architecture introduced by \cite{alex_jann}. To summarize, they consist of three modules. In the first module, a temporal convolution compresses the input sequence in time to a smaller number of points, called keypoints, defined by the ratio of original sequence length to the number of keypoints $r_\text{int}$. Here, the kernel size $k$ doubles from one convolutional layer to the next. In the second module, the compressed sequence gets processed by recurrently connected \acp{lif} units, resulting in an output that resembles the predicted trajectory but has the same length as the number of keypoints. Hence, in the final module, the downsampled sequence gets linearly interpolated back to the original sequence length. 

Since first presenting this network architecture in \cite{alex_jann}, we have tried different interpolation schemes, e.g., spline and higher-order interpolations. Nonetheless, they fail to provide considerable improvement. We argue that one instead has to understand the underlying motor control of the primates to invent an interpolation that resembles the movement between points of interest. Since linear interpolation with $r_\text{int}\leq8$ only results in an error of a few percentage points, we identify the improvement of the keypoint prediction itself as the primary goal.

In this work, we present three different models: a large uncompressed model with maximum achievable $R^2$ score, which we call BMnet (BenchMark net), and two versions of the realtime-capable model (see \ref{realtime}), one without compression and one with which we call RTnet and sRTnet ((small) RealTime net), respectively.

\subsection{Hyperparameter Search}
For the type of network explained above, there are four parameters defining a fixed architecture: Number of convolutional channels per layer $N_\text{C}$, kernel size $k$, number of \ac{lif} layers $N_\text{L}$, and number of \ac{lif} units per layer $N_\text{LIF}$. Additionally, the sequence length $L_\text{seq}$ defines how many consecutive numbers of data points the network is trained on. Note that $L_\text{seq}$ only affects training and validation. During testing, the network is instead inferred on a continuous data stream. The results of the searches for those hyperparameters are displayed in Figure \ref{fig:hparams_search}. Since we found that the L2 file consistently yielded the lowest accuracies and because hyperparameter searches take much time, the search was only conducted on that particular file using Optuna \cite{optuna}. 

While there seem to be optima for $N_\text{C}$, and $N_\text{LIF}$, the $R^2$ score does not stop improving when increasing $k$, $N_\text{L}$, and $L_\text{seq}$ across the considered value ranges. Since increasing them further comes at a computational cost, we limited them to the shown range and chose the maximum values as the optima. Moreover, the values are found by optimizing the $R^2$ score and might interfere with secondary objectives of the different models, e.g., resourcefulness or realtime capability. Hence, we argue that there are different optimums for the hyperparameters depending on the the model type. Table \ref{tab:hparams_values} summarizes the findings for optimal hyperparameter values of all three models.
\begin{figure}
    \centering
    \begin{tikzpicture}
    \begin{axis}[
        name=plot1,
        height=\h,
        width=\w,
        ytick={0.4, 0.6},
        xtick={0, 200},
        ylabel=$R^2$,
        xlabel=$N_\text{C}$,
    ]
        \addplot[black, only marks, mark size=\marsiz, mark=\mar] table {data/HPS/conv_channel.tex};
    \end{axis}
    \begin{axis}[
        name=plot2,
        at={($(plot1.east)+(\hd,0)$)},
        anchor=west,
        height=\h,
        width=\w,
        xtick={0, 20},
        yticklabels=\empty,
        xlabel=$k$,
    ]
        \addplot[black, only marks, mark size=\marsiz, mark=\mar] table {data/HPS/kernel_size.tex};
    \end{axis}
    \begin{axis}[
        name=plot3,
        at={($(plot2.east)+(\hd,0)$)},
        anchor=west,
        height=\h,
        width=\w,
        xtick={1, 2},
        yticklabels=\empty,
        xlabel=$N_\text{L}$,
        xmin=0.4,
        xmax=2.6,
    ]
        \addplot[black, only marks, mark size=\marsiz, mark=\mar] table {data/HPS/n_lif_layer.tex};
    \end{axis}
    \begin{axis}[
        name=plot4,
        at={($(plot3.east)+(\hd,0)$)},
        anchor=west,
        height=\h,
        width=\w,
        xtick={0, 100},
        ytick={0.4, 0.6},
        yticklabels=\empty,
        xlabel=$N_\text{LIF}$,
    ]
        \addplot[black, only marks, mark size=\marsiz, mark=\mar] table {data/HPS/n_lif.tex};
    \end{axis}
    \begin{axis}[
        name=plot5,
        at={($(plot4.east)+(\hd,0)$)},
        anchor=west,
        height=\h,
        width=\w,
        xtick={300, 3000},
        yticklabels=\empty,
        xlabel=$L_\text{seq}$,
    ]
        \addplot[black, only marks, mark size=\marsiz, mark=\mar] table {data/HPS/seq_len.tex};
    \end{axis}
\end{tikzpicture}
    \caption{Visualization of the hyperparameter search.}
    \label{fig:hparams_search}
\end{figure}
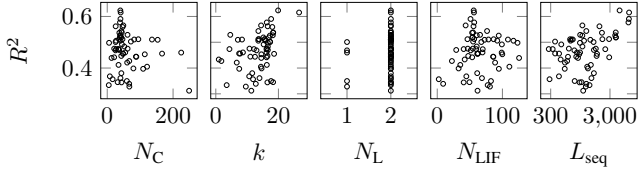
\begin{table}[]
    \centering
    \caption{Optimal hyperparameter values for all three considered models. The fixed point format is given in \textit{number of integer bits - number of fractional bits}.}

\begin{tabular}{lcccccccc}
\toprule
\textbf{Model} & \textbf{$N_\text{C}$} & \textbf{$k$} & \textbf{$N_\text{L}$} & \textbf{$N_\text{LIF}$} & \textbf{$L_\text{seq}$} & \textbf{$r_{\text{int}}$} & \makecell{\textbf{Weight}\\\textbf{Quantization}} & \makecell{\textbf{Buffer}\\\textbf{Quantization}} \\
\midrule

BMnet &40  &31  &2  &56  &4096 &8 &16\,float &16\,float\\
RTnet &40  &9  &2  &56  &4096 &4  &16\,float &16\,float\\
sRTnet &40  &9  &2  &56  &4096 &4  &1-7\,fixed &1-4\,fixed\\

\bottomrule

\end{tabular}

    \label{tab:hparams_values}
\end{table}

\subsection{Real-time Implementation}
\label{realtime}
Following the arguments presented in \cite{alex_jann}, we explore an iterative implementation architecture introduced in \ref{sec:architecture} for maximizing the real-time applicability of the networks.

\begin{figure}[t!]
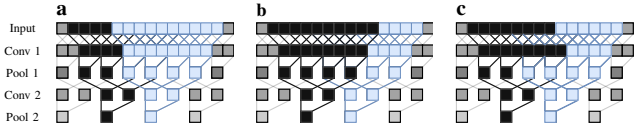

    \centering
    \includestandalone[width=1\linewidth]{figures/tikz/receptive_field_vis}
    \caption{Visualizations of the different buffer sizes (blue): \textbf{a} displays the buffer size per keypoint. \textbf{b} displays the buffer size for a new data update. \textbf{c} displays the buffers required to process new incoming data.}
    \label{fig:rec-field}
\end{figure}

For this, we compute the effective receptive field size in each convolutional and pooling layer required to compute one keypoint for each model (see Figure \ref{fig:rec-field}\textbf{a}). This is accomplished by working back from one keypoint and iteratively determining the size of the previous layer required to compute the output of a size represented by the receptive field of the current layer (see Algorithms \ref{alg:main_receptive_field}, \ref{alg:compute_buffer_size_keypoints}). The size determined by half of the buffer size of the initial layer + 1 step represents the model's latency since the original model's training is performed with centered kernels. 

In order to avoid repeating computations that have already been performed, for each layer's buffer, we determine the size of the new data required to add to the current receptive field that allows the computation of a new keypoint (see Figure \ref{fig:rec-field}\textbf{b}, Algorithms \ref{alg:main_receptive_field}, \ref{alg:compute_buffer_size_new_data_update}). The initial layer's \textit{Buffer Size New Data Update} determines the execution rate.

We then compute the size of the buffer needed to compute the new data in the next layer (see Figure \ref{fig:rec-field}\textbf{c}, Algorithms \ref{alg:main_receptive_field}, \ref{alg:compute_buffer_size_new_data}). These buffer sizes are used to implement the real-time version of the model.

\subsection{Hyperparameter selection for real-time applications}
\label{methods:rt}
\subsubsection{Execution Rate}
The execution rate is only determined by the interpolation size and, thus, by the number of convolutional/pooling layers. As such, an interpolation size of 8 steps implemented with three convolutional/pooling layers with 4ms per step offers an execution rate of 31.25 Hz. Similarly, an interpolation size of 4 steps implemented with two convolutional/pooling layers leads to an execution rate of 62.5 Hz.   

\subsubsection{Latency}
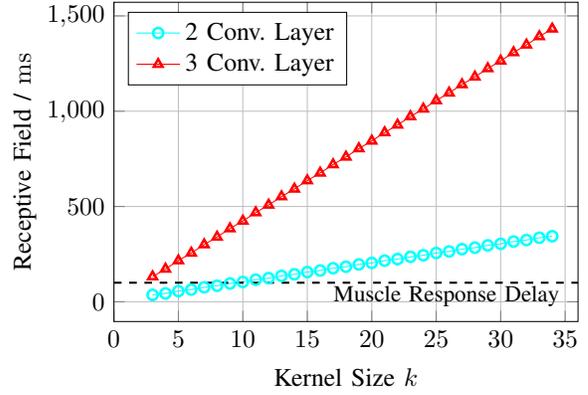
\begin{figure}
    \centering
    \begin{tikzpicture}
    \begin{axis}[
        height=6cm,
        width=8cm,
        xlabel={Kernel Size $k$},
        ylabel={Receptive Field / \si{\milli\second}},
        legend pos=north west,
        grid=major,
        xmin=0,
        xmax=36,
    ]

    \addplot[
        color=clr1,
        mark=o,
        mark size=3pt,
        mark options={mark size=2, line width=1pt}
    ] table [x=kernel_size, y=receptive_field_ms_2layer, col sep=comma]{data/receptive_fields.csv};
    \addlegendentry{2 Conv. Layer}
    \addplot[
        color=red,
        mark=triangle,
        mark size=3pt,
        mark options={mark size=2, line width=1pt}
    ] table [x=kernel_size, y=receptive_field_ms_3layer, col sep=comma]{data/receptive_fields.csv};
    \addlegendentry{3 Conv. Layer}
    \addplot[black, dashed, style=thick] coordinates {(0, 100) (36, 100)};

    \end{axis}

    \node[] at (4.6,0.35) {\small{Muscle Response Delay}};
\end{tikzpicture}
    \caption{Receptive field of the model architecture used in this work versus the kernel size $k$ of the convolutional blocks. It is important to note that the implementation of
our models doubles the size of the convolutional kernel in
each layer.}
    \label{fig:receptivefield_vs_kernelsize}
\end{figure}

Using the abovementioned algorithms, we compute the latency determined by different kernel size configurations in either two or three layers, which is plotted in Figure \ref{fig:receptivefield_vs_kernelsize}. It is important to note that the implementation of our models doubles the size of the convolutional kernel in each layer. Using this, we determine the required sizes of the kernels for a real-time oriented model, such that the latency falls under 100ms. In neuroscience literature, this is reported as the delay between brain impulse and voluntary muscle contraction \cite{kurtzer_long-latency_2015}. Thus, by using two convolutional layers and kernel sizes of 9 and 18, we can obtain a latency of 96\si{\milli\second}.

\subsection{Compression Methods}
The chosen hyperparameters for different compression methods are listed in Table \ref{tab:hparams_values}. Regularization parameter values were found using automized co-optimization of $R^2$, memory footprint, and AC operations as introduced in \cite{metric_paper}.



\textit{Spike Regularization}:
We utilize spike regularization during training to penalize low sparsity. We do this by adding the total sum multiplied with a regularization parameter $\lambda_S = 2.87 \times 10^{-3}$ to the total loss term.

\textit{Weight Regularization and Quantization}:
Quantization is very impactful when reducing the model footprint. In addition, it is incredibly effective to combine quantization with weight regularization, as more weights are quantized as zero when decreasing the bit width. The network is penalized for large weight values by adding the total sum of all absolute weight values to the loss function after scaling it with a regularization parameter $\lambda_\text{W}=1.71 \times 10^{-6}$. We differentiate between weight and buffer quantization, as this might have a different influence on the accuracy and because the memory dedicated to buffers is much smaller. We performed a grid search to minimize the loss of the $R^2$ score and arrive at the optimal bit widths. Hence, we apply fixed-point quantization with one integer bit and seven fractional bits to trainable parameters and quantize the buffers of the sRTnet using one integer bit and four fractional bits. Both quantization schemes use an additional bit for the sign. The same weight quantization has been used in \cite{ini_challenge}.

\subsection{Pretraining}
Following the work of \cite{zenke_decoding}, we use all available recording sessions of the primate reaching task \cite{primate_reaching_dataset} to perform pretraining for the two individual primates, thus generating four checkpoints (for each monkey, we pretrain a small and a large network). Most of the available files for primate Indy are recorded from the M1 Motor Cortex. However, there are some recordings for which the somatosensory S1 cortex was also recorded. In that case, we only took the M1 information. For primate Loco, all recordings are performed on both S1 and M1 cortices; thus, we use all available data here. 
The data is first augmented by generating overlapping samples of length 4096 with a stride of 1. Training is then performed for 50 epochs with early stopping respective to validation loss using a batch size of 2048 and a learning rate of $10^{-4}$.
\section{Results and Discussion}
\subsection{Benchmarking}
The results of the training of BMnet are summarized in Table \ref{tab:results_large}. 
\begin{table}
    \centering
    \caption{Comparison of BMnet and RTnet to related works that focus solely on optimizing the $R^2$ score on the Primate Reaching task, including the current state of the art for \acp{snn} and \acp{ann}.}

\begin{tabular}{lcc}
\toprule
\textbf{Model} & \textbf{Event-Based} & \textbf{Test $R^2$}\\

\midrule
Yik et al. - SNN3 \cite{neurobench} & \cmark & 0.633 \\
Liu et al. - bigRSNN \cite{zenke_decoding} & \cmark & $0.698\pm0.002$ \\
Vasilache et al. - LIF-t1 \cite{alex_jann} & \cmark & $0.648\pm0.022$ \\
Wang et al. \cite{ini_challenge} & \xmark & $0.710\pm0.050$ \\
Yik et al. - ANN3 \cite{neurobench} & \xmark & 0.615 \\
Vasilache et al. - GRU-t1 \cite{alex_jann} & \xmark & $0.707\pm0.012$ \\

\midrule
This Work - BMnet & \cmark & \textbf{0.717\,}$\pm$\textbf{\,0.004} \\
This Work - RTnet & \cmark & $0.685\pm0.006$ \\

\bottomrule

\end{tabular}

    \label{tab:results_large}
\end{table}
The obtained results for BMnet surpass the current benchmark on the Primate Reaching task for both \ac{snn} and \ac{ann}, making it the best-performing model yet while containing spiking activation functions. It should be noted that the results of the hyperparameter search hint at further improvement of these numbers when increasing $L_\text{seq}$ and $k$ during training. In this work, we intentionally put upper limits on those values to keep reasonable training times and model sizes. However, if small model size or realtime-applicability is not required, this leaves much room for improvement. Finally, increasing $L_\text{seq}$ only comes at the cost of more extended training time and hence should be leveraged by future work.

\subsection{Realtime Model}
\label{rt_discussion}
\begin{table*}
    \centering
    \caption{Comparison of sRTnet to related works that focus on co-optimizing the $R^2$ score, memory footprint, and computational demand on the Primate Reaching task, including the current state of the art for \acp{snn} and \acp{ann}. Additionally, the last column adresses the realtime capability of the respective networks (for details see \ref{rt_discussion}).}

\begin{tabular}{lcccccc}
\toprule
\textbf{Model} & \textbf{Event-Based} & \textbf{Test $R^2$} & \makecell{\textbf{Memory}\\\textbf{Footprint} / Bytes} & \textbf{MACs} & \textbf{ACs} & \makecell{\textbf{Realtime-}\\\textbf{Capable}}\\

\midrule
Yik et al. - SNN2 \cite{neurobench} & \cmark & 0.581 & 29\,248 & 0 & 414 & \cmark \\

Liu et al. - tinyRSNN \cite{zenke_decoding} & \cmark & 0.660$\pm$0.002 & \textbf{27\,144} & 0 & $304\pm12$ & \cmark  \\

Vasilache et al. - LIF-t2 \cite{alex_jann} & \cmark & $0.566\pm0.016$ & 168\,596 & $201\pm0$ & $254.0\pm0.8$ & \xmark \\

Yik et al. - ANN2 \cite{neurobench} & \xmark & 0.576 & 27\,160 & 4\,970 & 0 & \xmark \\

Vasilache et al. - GRU-t2 \cite{alex_jann} & \xmark & $0.621\pm0.014$ & 174\,104 & $627\pm0$ & $248\pm0$ & \xmark \\

\midrule
This Work - sRTnet & \cmark & \textbf{0.675}\,$\pm$\,\textbf{0.011} & 105\,269 & 12\,274\,$\pm$\,37 & $326\pm4$ & \cmark \\
\bottomrule

\end{tabular}

    \label{tab:results_small}
\end{table*}

Using the largest kernel size determined in the previous section that allows realtime deployment, we train the RTnet model to breach the gap between very small and very big networks while still keeping realtime deployment in mind (Table \ref{tab:results_large}). Compared with our BMnet, which uses 3 convolutional layers of sizes 31, 62, and 124 and has a receptive field of 652 points and latency of 1\,308\,ms, the RTnet has only 2 convolutional layers of size 9 and 18 with a receptive field of 46 points and 96\,ms latency. 

We apply our proposed compression methodology to the RTnet to obtain a smaller version, sRTnet (Table \ref{tab:results_small}). This allows us to reduce the footprint roughly by two thirds and increase connection sparsity to 15\% at the cost of 0.01 $R^2$. When comparing the sRTnet to the small models in the literature, we observe the highest reported $R^2$ score.

Comparing the memory footprint and number of MAC operations of sRTnet to our previous versions, we notice a substantial increase, which we attribute to the increase in the sizes of the convolutional kernels (from 3, 3 to 9, 18) and convolutional channels (10 to 40). These are also where sRTnet compares poorly with related works, especially tinyRSNN, considering that the difference in $R^2$ is only a few percentage points. Hence, we conclude that next iterations should decrease footprint and MACs by cutting down on the large convolutions while keeping the improvements in accuracy due to the large temporal context window of the network. Additionally, applying pruning to further reduce the memory footprint is desirable but has presented a challenge. When incrementally increasing the global unstructured pruning amount, we very quickly observed a sharp drop in accuracy. Hence, more sophisticated pruning methods should be considered for future work.
Finally, sparsely connected networks benefit from event-based updating of the neuronal units \cite{jann_sven}. Hence, such event-based neuron models should also be utilized in next-generation models. In terms of ACs, sRTnet displays a computational load comparable to that of related works. 

In \cite{zenke_decoding}, the authors compare ACs and MACs based on their relative energy cost. By that standard, the combined computational demand of sRTnet is far larger than that of the other networks. However, we argue, that whether the improvements in $R^2$ are worth the increased footprint and computational load heavily depends on the target platform the system will be deployed on. As sRTnet is a hybrid network with separable convolutional and spiking sub-nets, it is an optimal candidate for deployment on a hybrid computing platform consisting of neurormophic and CNN accelerators. Finally, deploying all networks on their respective optimal target hardware platform would enable a truly fair performance comparison at runtime.

\subsection{Realtime-Capability of Networks}
\label{rt_discussion}
To determine the realtime capability of all networks, we analyzed their computing principles and derived values for latency and execution rate which are mostly determined by input buffering and the temporal density of the outputs, e.g., whether the output is interpolated or not. The exact numbers and whether they can be considered realtime-capable is summarized in Table \ref{tab:rate_delay} in the supplementaries. As written in \ref{methods:rt}, the upper latency limit is around 100\,ms to be considered non-noticeable by patients. As for execution rates, current research in \acp{bmi} argues that having small latency is far more important than having high control rates. However, motor control prostheses commonly are controlled at rates of around 10\,Hz and can even benefit from slightly higher rates \cite{refresh_rate_bmi}. Based on this, we declare a network to be realtime-capable if it has a maximum delay of 100\,ms and a minimum execution rate of 10\,Hz.

\subsection{Synergy Between Motor and Somatosensory Cortex Data}
\begin{figure}
    \centering
    \begin{tikzpicture}
    \begin{axis}[
        ybar,
        bar width=.3cm,
        width=9cm,
        height=5cm,
        enlarge x limits=0.25,
        ylabel={Test $R^2$},
        symbolic x coords={L1 file, L2 file, L3 file},
        xtick=data,
        ymin=0,
        legend style={at={(0.5,1.25)},
            anchor=north,legend columns=-1},
        enlarge y limits={value=.1,upper},
        ]
        \addplot+[error bars/.cd, y dir=both, y explicit] coordinates {
            (L1 file,0.6169) +- (0,0.0080)
            (L2 file,0.5816) +- (0,0.0111)
            (L3 file,0.6195) +- (0,0.0103)
            };
        \addplot+[error bars/.cd, y dir=both, y explicit] coordinates {
            (L1 file,0.3863) +- (0,0.0148)
            (L2 file,0.3111) +- (0,0.0702)
            (L3 file,0.5712) +- (0,0.0092)
            };
        \addplot+[error bars/.cd, y dir=both, y explicit] coordinates {
            (L1 file,0.6394) +- (0,0.0150)
            (L2 file,0.5215) +- (0,0.0423)
            (L3 file,0.6966) +- (0,0.0044)
            };
        \addplot+[error bars/.cd, y dir=both, y explicit] coordinates {
            (L1 file,0.6187) +- (0,0.0270)
            (L2 file,0.5698) +- (0,0.0295)
            (L3 file,0.6770) +- (0,0.0186)
            };
        \addplot+[error bars/.cd, y dir=both, y explicit] coordinates {
            (L1 file,0.6376) +- (0,0.0260)
            (L2 file,0.5460) +- (0,0.0373)
            (L3 file,0.6555) +- (0,0.0252)
            };
        \legend{M1, S1, M1+S1, pre-LIF, post-LIF}
    \end{axis}
\end{tikzpicture}
    \caption{Summary of accuracy scores for different processing approaches of the motor and somatosensory cortex spike trains. The error bars define the standard deviation of the mean $R^2$ value.}
    \label{fig:m1s1_barplot}
\end{figure}
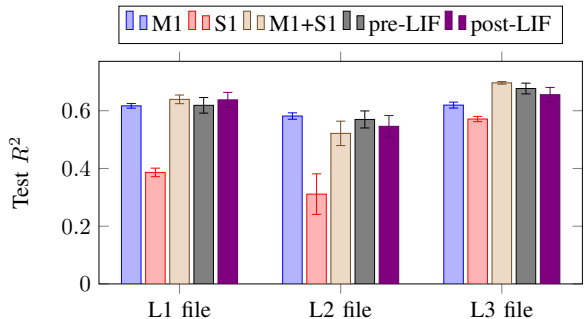

As mentioned above, the three Loco recordings consist of M1 and S1 data. Since most files consist only of M1 recordings, we assume that the S1 data plays a supplementary role in decoding the arm velocities. Assuming the benefit of adding them in terms of $R^2$ is little to none, one can also discard them and save considerable computational resources. On the other hand, if the decoding benefits from additional S1 recordings, separately processing M1 and S1 data might make sense. To test this, we train BMnet networks on only M1, only S1, and both M1 and S1 recordings for files L1 to L3. Additionally, to test the separate processing of both brain recording types, we train networks that have separate convolutions (pre-LIF) and networks that have separate convolutions and layers of \ac{lif} units (post-LIF) before merging the predictions of the M1 and S1 spikes. Figure \ref{fig:m1s1_barplot} summarizes the resulting accuracies.

A clear performance gap exists between networks trained on M1 data and networks trained on S1 data only. Interestingly, this gap is relatively small for the L3 recording. Hence, good training data should always contain M1 spikes. Moreover, processing M1 and S1 data together results in better $R^2$ scores than only training on M1 data. This is not entirely true for the L2 file. Looking at the error bars of such networks that are trained on data including the S1 spikes of the L2 file, we argue that the quality of the S1 data might not be optimal, which results in a deterioration of the accuracies. The error bars further hint that the single best trainings were always performed on a combination of M1 and S1 data. Finally, our approach of separately processing M1 and S1 spikes yielded no performance improvement. We argue that due to the different nature of the motor and somatosensory cortices, it is probable that there still exists a promising method of separately processing their spikes. We conclude that training networks on both M1 and S1 is beneficial, even at the cost of a larger footprint and higher computational demand. Still, when confronted with very small memory requirements, training only on M1 data might be the better option.

\section{Conclusion and Outlook}
\label{sec:conclusion}
We presented the new iteration of our \ac{snn}-based neural decoder for reconstructing finger velocities from cortical spike trains \cite{alex_jann}. We first benchmarked the model via hyperparameter optimization on the Primate Reaching dataset, improving on the $R^2$ of our previous results by more than 7\% and beating the current state of the art of \acp{ann} and \acp{snn} by 1\% and 1.9\%, respectively \cite{ini_challenge, zenke_decoding}. 

Meanwhile, our compressed models also improved the state of the art by an $R^2$ score of 1.5\% \cite{zenke_decoding}. However, they displayed increased memory footprints by a factor of four. The number of accumulate operations is in the same range and the multiply-and-accumulate operations are one to two orders of magnitudes larger when compared to related works. The latter of which is a downside of the increased kernel sizes that are introduced for more accurate decoding. We argue that slight disadvantages in memory and computing resources can be overcome by more sophisticated pruning and event-driven updating, as this would further decrease the number of operations and non-zero weights. 

Furthermore, we identified the shortcomings of our models in terms of latency and introduced the necessary adaptations to their architecture to become realtime-capable. We then discussed the implications of that realtime architecture and adjusted hyperparameters so that the \ac{snn} can be inferred seemingly latency-free. 

Future work will focus on implementing the networks presented in this work on neuromorphic hardware, e.g., Loihi2 \cite{loihi2} and hybrid computing platforms including neuromorphic accelerators. This will test the real-world applicability of our approach while taking full advantage of the ability of neuromorphic systems to run on low power. By doing so, we ultimately address the requirements of wireless \acp{ibmi} regarding heat dissipation, bandwidth limitation, and battery lifetime \cite{wireless_ibmi, wireless_ibmi2, thermal_dissipation}. With this, we step closer to the successful demonstration of \acp{snn} and neuromorphic hardware as promising candidates for leading-edge neuroprosthetics with the possibility to change the lives of millions of people with paralysis worldwide.

\bibliographystyle{ieeetr}
\bibliography{refs}

\section{Supplementary Information}

\begin{algorithm}[H]
\caption{BSizeNewData}
\label{alg:compute_buffer_size_new_data}
\begin{algorithmic}[1]
\Function{BSizeNewData}{$\mathbf{B}_{\text{new\_data\_update}}, \mathbf{K}_{\text{conv}}$}
    \State $\mathbf{B}_{\text{new\_data}} \gets []$
    \For{$i = 0$ to \text{length}($\mathbf{B}_{\text{new\_data\_update}}$) - 1}
        \If{$i$ is even}
            \State $B \gets B_{\text{new\_data\_update}, i} - 1 + K_{\text{conv}, i/2}$
            \State Append $B$ to $\mathbf{B}_{\text{new\_data}}$
        \Else
            \State Append $B_{\text{new\_data\_update}, i}$ to $\mathbf{B}_{\text{new\_data}}$
        \EndIf
    \EndFor
    \State \Return $\mathbf{B}_{\text{new\_data}}$
\EndFunction
\end{algorithmic}
\end{algorithm}

\begin{algorithm}[H]
\caption{Buffer Sizes Calculation}
\label{alg:main_receptive_field}
\begin{algorithmic}[1]
\Require Number of layers $N_{\text{conv}}$, $N_{\text{pool}}$
\Require Convolution kernel sizes $\mathbf{K}_{\text{conv}}$, strides $\mathbf{S}_{\text{conv}}$
\Require Pooling kernel sizes $\mathbf{K}_{\text{pool}}$, strides $\mathbf{S}_{\text{pool}}$
\Ensure buffer size keypoints, buffer size new data, buffer size new data update

\State Initialize $R \gets 1$, $B_{\text{update}} \gets 1$, $\mathbf{R}_{\text{list}} \gets []$, $\mathbf{B}_{\text{update\_list}} \gets []$
\For{$i = 0$ to $N_{\text{conv}} + N_{\text{pool}} - 1$}
    \If{$i$ is even} \Comment{Convolutional layer}
        \State $R \gets (K_{\text{conv}, i/2} - 1) \cdot B_{\text{update}} + R$
        \State $B_{\text{update}} \gets B_{\text{update}} \cdot S_{\text{conv}, i/2}$
    \Else \Comment{Pooling layer}
        \State $R \gets (K_{\text{pool}, (i-1)/2} - 1) \cdot B_{\text{update}} + R$
        \State $B_{\text{update}} \gets B_{\text{update}} \cdot S_{\text{pool}, (i-1)/2}$
    \EndIf
    \State Append $R$ to $\mathbf{R}_{\text{list}}$, $B_{\text{update}}$ to $\mathbf{B}_{\text{update\_list}}$
\EndFor

\State $\mathbf{B}_{\text{keypoints}} \gets \Call{BSizeKeypoints}{R, \mathbf{K}_{\text{conv}}, \mathbf{S}_{\text{pool}}}$
\State $\mathbf{B}_{\text{new\_data\_update}} \gets \Call{BSizeNewDataUpdate}{\mathbf{B}_{\text{update\_list}}}$
\State $\mathbf{B}_{\text{new\_data}} \gets \Call{BSizeNewData}{\mathbf{B}_{\text{new\_data\_update}}, \mathbf{K}_{\text{conv}}}$
\State \Return $\mathbf{B}_{\text{keypoints}}, \mathbf{B}_{\text{new\_data}}, \mathbf{B}_{\text{new\_data\_update}}$
\end{algorithmic}
\end{algorithm}

\begin{algorithm}[H]
\caption{BSizeNewDataUpdate}
\label{alg:compute_buffer_size_new_data_update}
\begin{algorithmic}[1]
\Function{BSizeNewDataUpdate}{$\mathbf{B}_{\text{update\_list}}$}
    \State $\mathbf{B}_{\text{update}} \gets \mathbf{B}_{\text{update\_list}}$
    \State Append last element of $\mathbf{B}_{\text{update}}$ to itself
    \State Reverse $\mathbf{B}_{\text{update}}$ and remove its last element
    \State \Return $\mathbf{B}_{\text{update}}$
\EndFunction
\end{algorithmic}
\end{algorithm}

\begin{algorithm}[H]
\caption{BSizeKeypoints}
\label{alg:compute_buffer_size_keypoints}
\begin{algorithmic}[1]
\Function{BSizeKeypoints}{$R, \mathbf{K}_{\text{conv}}, \mathbf{S}_{\text{pool}}$}
    \State $C \gets R$, $\mathbf{B}_{\text{keypoints}} \gets []$
    \For{$i = 0$ to $2 \cdot \text{length}(\mathbf{K}_{\text{conv}}) - 1$}
        \If{$i$ is even}
            \State Append $C$ to $\mathbf{B}_{\text{keypoints}}$
        \Else
            \State $C \gets C - K_{\text{conv}, i/2} + 1$
            \State Append $C$ to $\mathbf{B}_{\text{keypoints}}$
            \State $C \gets C / S_{\text{pool}, i/2}$
        \EndIf
    \EndFor
    \State \Return $\mathbf{B}_{\text{keypoints}}$
\EndFunction
\end{algorithmic}
\end{algorithm}



\begin{table}[H]
    \centering
    \caption{Comparison of sRTnet to related works that co-optimize the computational demand and $R^2$ score on the Primate Reaching task regarding latency and refresh rate. From those values we derive wether the network is realtime-capable. For details see \ref{rt_discussion}.}

\begin{tabular}{lccc}
\toprule
\textbf{Model} & \makecell{\textbf{Latency}\\/ ms} & \makecell{\textbf{Execution}\\\textbf{Rate} / Hz} & \makecell{\textbf{Realtime-}\\\textbf{Capable}}\\

\midrule
Yik et al. - SNN2 \cite{neurobench} & \textbf{4} & \textbf{250} & \cmark\\

Liu et al. - tinyRSNN \cite{zenke_decoding} & \textbf{4} & \textbf{250} & \cmark\\

Vasilache et al. - LIF-t2 \cite{alex_jann} & 4\,096 & 0.244 & \xmark\\

Yik et al. - ANN2 \cite{neurobench} & 200 & \textbf{250} & \xmark\\

Vasilache et al. - GRU-t2 \cite{alex_jann} & 4\,096 & 0.244 & \xmark\\

\midrule
This Work - sRTnet & 96 & 62.5 & \cmark\\
\bottomrule

\end{tabular}

    \label{tab:rate_delay}
\end{table}

\end{document}